\documentclass{article} 
\usepackage{iclr2016_conference,times}
\usepackage{hyperref}
\usepackage{url}

\usepackage{amsmath,amssymb,amsthm,pgf,caption,subcaption,placeins,listings,multicol}
\usepackage{macros}

\title{SparkNet: Training Deep Networks in Spark}

\author{Philipp Moritz\thanks{Both authors contributed equally.},\,\, Robert Nishihara${}^*$\hspace{-4.4pt},\,\, Ion Stoica,\, Michael I.~Jordan\\
Electrical Engineering and Computer Science\\
University of California\\
Berkeley, CA 94720, USA \\
\texttt{\{pcmoritz,rkn,istoica,jordan\}@eecs.berkeley.edu} \\
}

\iclrfinalcopy 

\begin{document}

\maketitle

\begin{abstract}
Training deep networks is a time-consuming process, with networks for object recognition often requiring multiple days to train. 
For this reason, leveraging the resources of a cluster to speed up training is an important area of work. 
However, widely-popular batch-processing computational frameworks like MapReduce and Spark were not designed to support the asynchronous and communication-intensive workloads of existing distributed deep learning systems. 
We introduce SparkNet, a framework for training deep networks in Spark. 
Our implementation includes a convenient interface for reading data from Spark RDDs, a Scala interface to the Caffe deep learning framework, and a lightweight multi-dimensional tensor library. 
Using a simple parallelization scheme for stochastic gradient descent, SparkNet scales well with the cluster size and tolerates very high-latency communication. 
Furthermore, it is easy to deploy and use with no parameter tuning, and it is compatible with existing Caffe models. 
We quantify the dependence of the speedup obtained by SparkNet on the number of machines, the communication frequency, and the cluster's communication overhead, and we benchmark our system's performance on the ImageNet dataset. 
\end{abstract}

\section{Introduction}

Deep learning has advanced the state of the art in a number of application domains. 
Many of the recent advances involve fitting large models (often several hundreds megabytes) to larger datasets (often hundreds of gigabytes). 
Given the scale of these optimization problems, training can be time-consuming, often requiring multiple days on a single GPU using stochastic gradient descent (SGD). 
For this reason, much effort has been devoted to leveraging the computational resources of a cluster to speed up the training of deep networks (and more generally to perform distributed optimization). 

Many attempts to speed up the training of deep networks rely on asynchronous, lock-free optimization \citep{dean2012large,chilimbi2014project}. 
This paradigm uses the parameter server model \citep{li2014scaling,ho2013more}, in which one or more master nodes hold the latest model parameters in memory and serve them to worker nodes upon request.
The nodes then compute gradients with respect to these parameters on a minibatch drawn from the local data shard.
These gradients are shipped back to the server, which updates the model parameters.

At the same time, batch-processing frameworks enjoy widespread usage and have been gaining in popularity. 
Beginning with MapReduce \citep{dean2008mapreduce}, a number of frameworks for distributed computing have emerged to make it easier to write distributed programs that leverage the resources of a cluster \citep{zaharia2010spark,isard2007dryad,murray2013naiad}. 
These frameworks have greatly simplified many large-scale data analytics tasks. 
However, state-of-the-art deep learning systems rely on custom implementations to facilitate their asynchronous, communication-intensive workloads. 
One reason is that popular batch-processing frameworks \citep{dean2008mapreduce,zaharia2010spark} are not designed to support the workloads of existing deep learning systems. 
SparkNet implements a scalable, distributed algorithm for training deep networks that lends itself to batch computational frameworks such as MapReduce and Spark and works well out-of-the-box in bandwidth-limited environments. 

The benefits of integrating model training with existing batch frameworks are numerous. 
Much of the difficulty of applying machine learning has to do with obtaining, cleaning, and processing data as well as deploying models and serving predictions. 
For this reason, it is convenient to integrate model training with the existing data-processing pipelines that have been engineered in today's distributed computational environments. 
Furthermore, this approach allows data to be kept in memory from start to finish, whereas a segmented approach requires writing to disk between operations. 
If a user wishes to train a deep network on the output of a SQL query or on the output of a graph computation and to feed the resulting predictions into a distributed visualization tool, this can be done conveniently within a single computational framework. 

We emphasize that the hardware requirements of our approach are minimal. 
Whereas many approaches to the distributed training of deep networks involve heavy communication (often communicating multiple gradient vectors for every minibatch), our approach gracefully handles the bandwidth-limited setting while also taking advantage of clusters with low-latency communication. 
For this reason, we can easily deploy our algorithm on clusters that are not optimized for communication. 
Our implementation works well out-of-the box on a five-node EC2 cluster in which broadcasting and collecting model parameters (several hundred megabytes per worker) takes on the order of $20$ seconds, and performing a single minibatch gradient computation requires about $2$ seconds (for AlexNet).
We achieve this by providing a simple algorithm for parallelizing SGD that involves minimal communication and lends itself to straightforward implementation in batch computational frameworks. 
Our goal is not to outperform custom computational frameworks but rather to propose a system that can be easily implemented in popular batch frameworks and that performs nearly as well as what can be accomplished with specialized frameworks.

\section{Implementation}

\begin{figure}[t]
  \centering
  \includegraphics[scale=0.8]{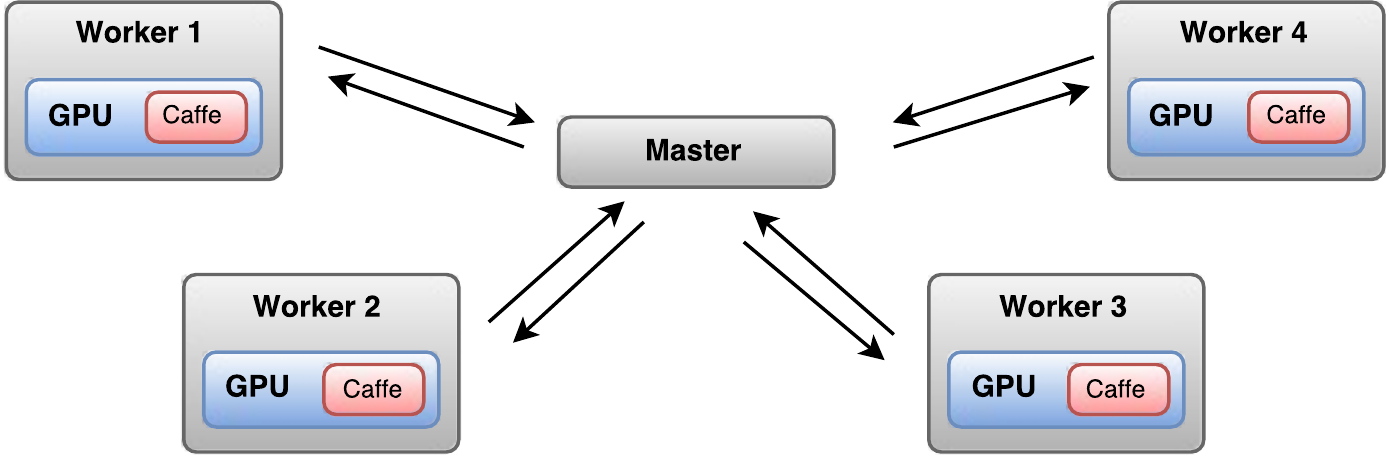}
  \caption{
    This figure depicts the SparkNet architecture.
  }
  \label{fig:sparknet_architecture}
\end{figure}

\begin{figure}[b]
  \lstset{basicstyle=\linespread{1.5}\small\ttfamily, keywords={class, def}, frame=tb,
    label=lst:SparkNet_API, captionpos=b,
    caption={SparkNet API}}
\begin{lstlisting}
class Net {
  def Net(netParams: NetParams): Net
  def setTrainingData(data: Iterator[(NDArray,Int)])
  def setValidationData(data: Iterator[(NDArray,Int)])
  def train(numSteps: Int)
  def test(numSteps: Int): Float
  def setWeights(weights: WeightCollection)
  def getWeights(): WeightCollection
}
\end{lstlisting}
\vspace{-1em}
\end{figure}

Here we describe our implementation of SparkNet. 
SparkNet builds on Apache Spark \citep{zaharia2010spark} and the Caffe deep learning library \citep{jia2014caffe}. In addition, we use Java Native Access for accessing Caffe data and weights natively from Scala, and we use the Java implementation of Google Protocol Buffers to allow the dynamic construction of Caffe networks at runtime.

\begin{figure}
  \lstset{basicstyle=\linespread{1.2}\small\ttfamily, keywords={class, def}, frame=tb,
    label=lst:lenet_specification, captionpos=b,
    caption={Example network specification in SparkNet}}
\begin{lstlisting}
val netParams = NetParams(
  RDDLayer("data", shape=List(batchsize, 1, 28, 28)),
  RDDLayer("label", shape=List(batchsize, 1)),
  ConvLayer("conv1", List("data"), kernel=(5,5), numFilters=20),
  PoolLayer("pool1", List("conv1"), pool=Max, kernel=(2,2), stride=(2,2)),
  ConvLayer("conv2", List("pool1"), kernel=(5,5), numFilters=50),
  PoolLayer("pool2", List("conv2"), pool=Max, kernel=(2,2), stride=(2,2)),
  LinearLayer("ip1", List("pool2"), numOutputs=500),
  ActivationLayer("relu1", List("ip1"), activation=ReLU),
  LinearLayer("ip2", List("relu1"), numOutputs=10),
  SoftmaxWithLoss("loss", List("ip2", "label"))
)
\end{lstlisting}
\vspace{-1em}
\end{figure}

The Net class wraps Caffe and exposes a simple API containing the methods shown in \lstref{lst:SparkNet_API}. 
The \texttt{NetParams} type specifies a network architecture, and the \texttt{WeightCollection} type is a map from layer names to lists of weights. It allows the manipulation of network components and the storage of weights and outputs for individual layers. 
To facilitate manipulation of data and weights without copying memory from Caffe, we implement the \texttt{NDArray} class, which is a lightweight multi-dimensional tensor library. 
One benefit of building on Caffe is that any existing Caffe model definition or solver file is automatically compatible with SparkNet. There is a large community developing Caffe models and extensions, and these can easily be used in SparkNet.
By building on top of Spark, we inherit the advantages of modern batch computational frameworks. 
These include the high-throughput loading and preprocessing of data and the ability to keep data in memory between operations. 
In \lstref{lst:lenet_specification}, we give an example of how network architectures can be specified in SparkNet. 
In addition, model specifications or weights can be loaded directly from Caffe files. 
An example sketch of code that uses our API to perform distributed training is given in \lstref{lst:training_example}.
\begin{figure}[t]
  \lstset{basicstyle=\linespread{1.2}\small\ttfamily, keywords={def}, frame=tb,
    label=lst:training_example, captionpos=b,
    caption={Distributed training example}}
\begin{lstlisting}
var trainData = loadData(...)
var trainData = preprocess(trainData).cache()
var nets = trainData.foreachPartition(data => {
  var net = Net(netParams)
  net.setTrainingData(data)
  net)
var weights = initialWeights(...)
for (i <- 1 to 1000) {
  var broadcastWeights = broadcast(weights)
  nets.map(net => net.setWeights(broadcastWeights.value))
  weights = nets.map(net => {
    net.train(50)
    net.getWeights()}).mean() // an average of WeightCollection objects
}
\end{lstlisting}
\vspace{-1em}
\end{figure}


\subsection{Parallelizing SGD} \label{sec:parallelizing_sgd}

To perform well in bandwidth-limited environments, we recommend a parallelization scheme for SGD that requires minimal communication. 
This approach is not specific to SGD. 
Indeed, SparkNet works out of the box with any Caffe solver. 

The parallelization scheme is described in \lstref{lst:training_example}. 
Spark consists of a single master node and a number of worker nodes. 
The data is split among the Spark workers. 
In every iteration, the Spark master broadcasts the model parameters to each worker. 
Each worker then runs SGD on the model with its subset of data for a fixed number of iterations $\tau$ (we use $\tau=50$ in \lstref{lst:training_example}) or for a fixed length of time, after which the resulting model parameters on each worker are sent to the master and averaged to form the new model parameters. 
We recommend initializing the network by running SGD for a small number of iterations on the master. 
A similar and more sophisticated approach to parallelizing SGD with minimal communication overhead is discussed in \citet{sixin2015deep}. 

The standard approach to parallelizing each gradient computation requires broadcasting and collecting model parameters (hundreds of megabytes per worker and gigabytes in total) after every SGD update, which occurs tens of thousands of times during training. 
On our EC2 cluster, each broadcast and collection takes about twenty seconds, putting a bound on the speedup that can be expected using this approach without better hardware or without partitioning models across machines. 
Our approach broadcasts and collects the parameters a factor of $\tau$ times less for the same number of iterations. 
In our experiments, we set $\tau=50$, but other values seem to work about as well. 

We note that Caffe supports parallelism across multiple GPUs within a single node. 
This is not a competing form of parallelism but rather a complementary one. 
In some of our experiments, we use Caffe to handle parallelism within a single node, and we use the parallelization scheme described in \lstref{lst:training_example} to handle parallelism across nodes. 

\section{Experiments}

In \secref{sec:training_benchmarks}, we will benchmark the performance of SparkNet and measure the speedup that our system obtains relative to training on a single node. 
However, the outcomes of those experiments depend on a number of different factors. 
In addition to $\tau$ (the number of iterations between synchronizations) and $K$ (the number of machines in our cluster), they depend on the communication overhead in our cluster $S$. 
In \secref{sec:theoretical_considerations}, we find it instructive to measure the speedup {\em in the idealized case of zero communication overhead} ($S=0$). 
This idealized model gives us an upper bound on the maximum speedup that we could hope to obtain in a real-world cluster, and it allows us to build a model for the speedup as a function of $S$ (the overhead is easily measured in practice). 

\subsection{Theoretical Considerations} \label{sec:theoretical_considerations}

Before benchmarking our system, we determine the maximum possible speedup that could be obtained in principle in a cluster with no communication overhead. 
We determine the dependence of this speedup on the parameters $\tau$ (the number of iterations between synchronizations) and $K$ (the number of machines in our cluster). 

\subsubsection{Limitations of Naive Parallelization}
\begin{figure}[t]
\centering
\begin{subfigure}{0.9\textwidth}
  \centering
  \includegraphics[scale=0.7]{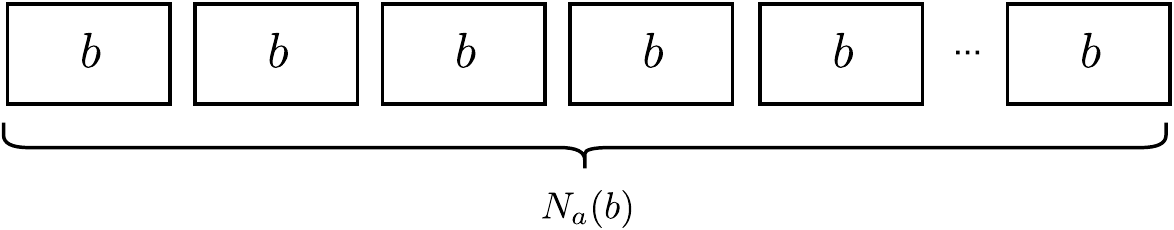}
  \caption{
    This figure depicts a serial run of SGD. 
    Each block corresponds to a single SGD update with batch size $b$. 
    The quantity $N_a(b)$ is the number of iterations required to achieve an accuracy of $a$. 
  }
  \label{fig:sgd_flow}
\end{subfigure}

\vspace{10pt}

\begin{subfigure}{0.9\textwidth}
  \centering
  \includegraphics[scale=0.7]{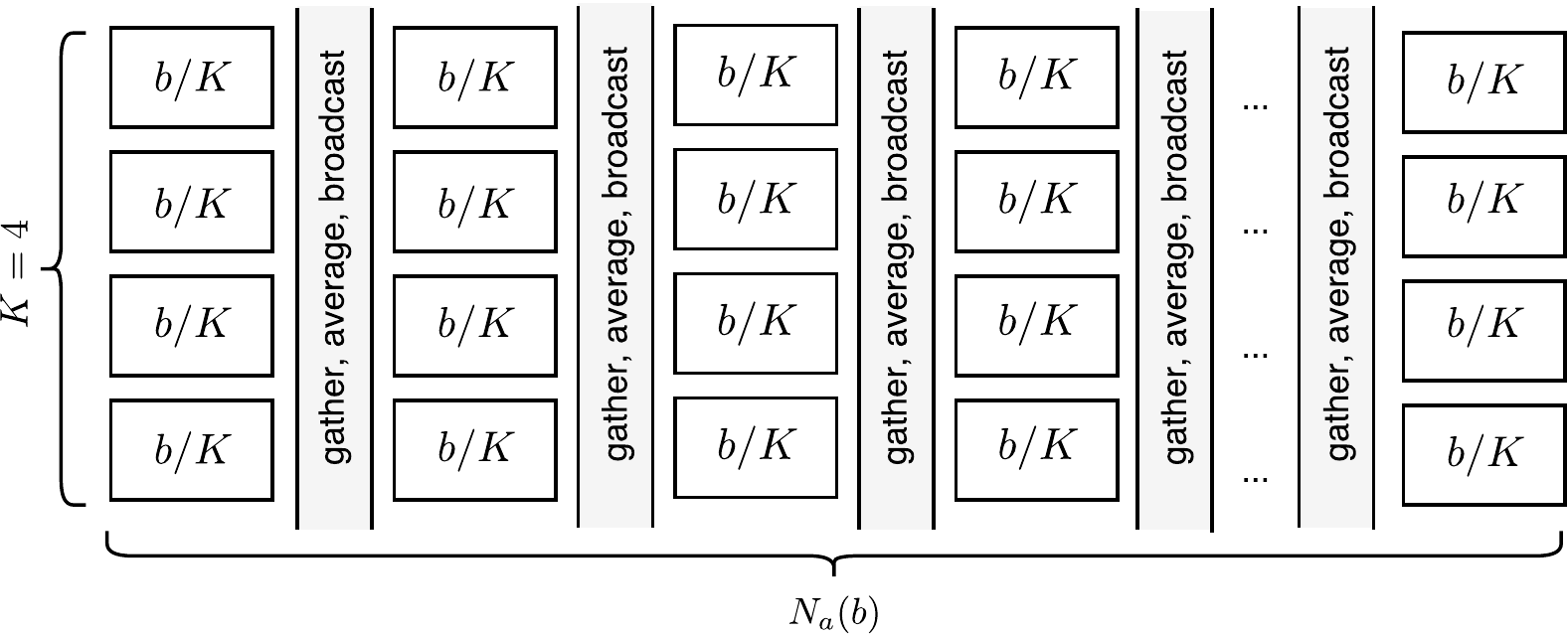}
  \caption{
    This figure depicts a parallel run of SGD on $K=4$ machines under a naive parallelization scheme. 
    At each iteration, each batch of size $b$ is divided among the $K$ machines, the gradients over the subsets are computed separately on each machine, the updates are aggregated, and the new model is broadcast to the workers. 
    Algorithmically, this approach is exactly equivalent to the serial run of SGD in \figref{fig:sgd_flow} and so the number of iterations required to achieve an accuracy of $a$ is the same value $N_a(b)$. 
  }
  \label{fig:naive_flow}
\end{subfigure}

\vspace{10pt}

\begin{subfigure}{0.9\textwidth}
  \centering
  \includegraphics[scale=0.7]{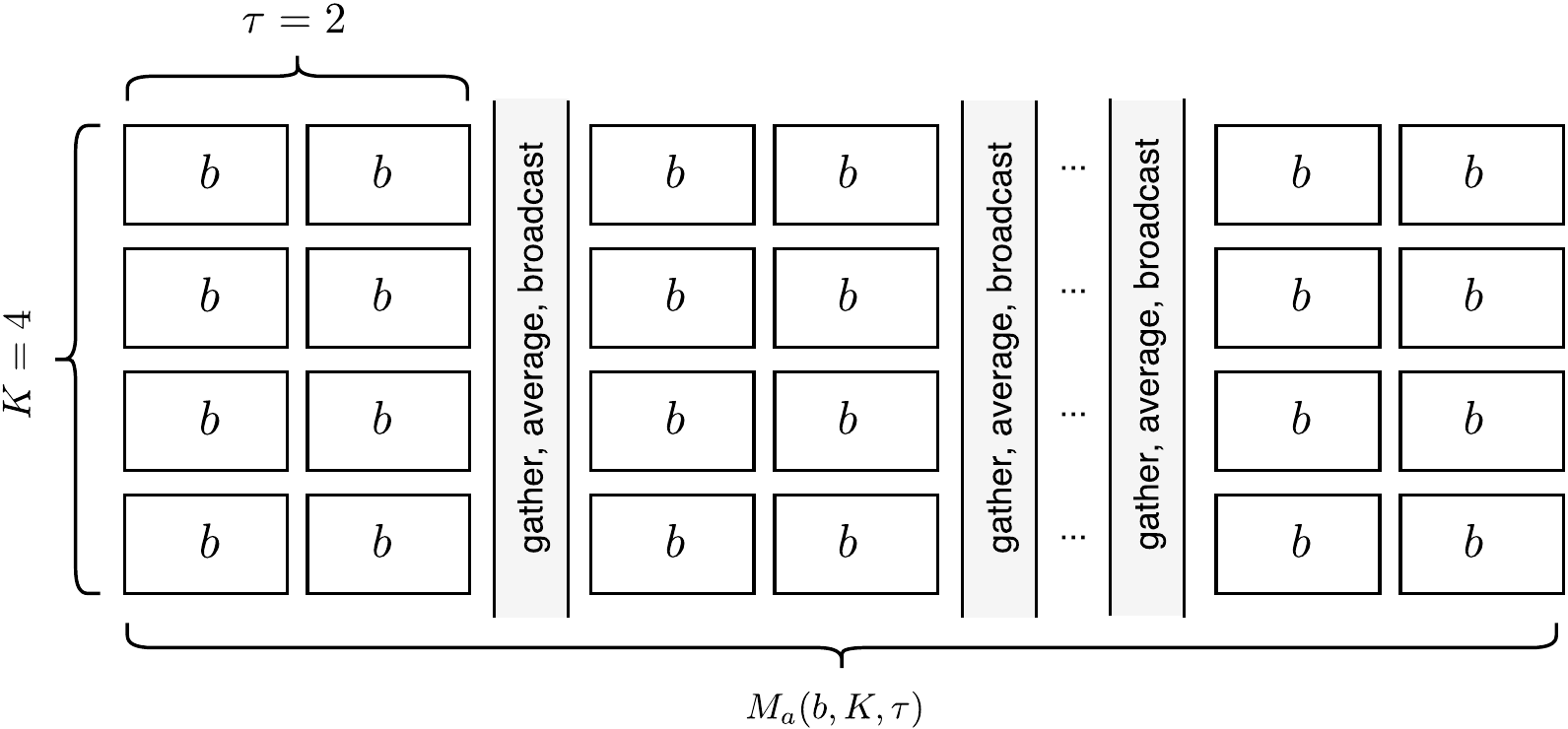}
  \caption{
    This figure depicts a parallel run of SGD on $K=4$ machines under SparkNet's parallelization scheme. 
    At each step, each machine runs SGD with batch size $b$ for $\tau$ iterations, after which the models are aggregated, averaged, and broadcast to the workers. 
    The quantity $M_a(b,K,\tau)$ is the number of rounds (of $\tau$ iterations) required to obtain an accuracy of $a$. 
    The total number of parallel iterations of SGD under SparkNet's parallelization scheme required to obtain an accuracy of $a$ is then $\tau M_a(b,K,\tau)$. 
  }
  \label{fig:sparknet_flow}
\end{subfigure}
\caption{Computational models for different parallelization schemes.}
\label{fig:flows}
\end{figure}

To begin with, we consider the theoretical limitations of a naive parallelism scheme which parallelizes SGD by distributing each minibatch computation over multiple machines (see \figref{fig:naive_flow}). 
Let $N_a(b)$ be the number of serial iterations of SGD required to obtain an accuracy of $a$ when training with a batch size of $b$ (when we say accuracy, we are referring to test accuracy). 
Suppose that computing the gradient over a batch of size $b$ requires $C(b)$ units of time. 
Then the running time required to achieve an accuracy of $a$ with serial training is
\begin{equation} \label{eq:naive_time}
  N_a(b)C(b). 
\end{equation}
A naive parallelization scheme attempts to distribute the computation at each iteration by dividing each minibatch between the $K$ machines, computing the gradients separately, and aggregating the results on one node. 
Under this scheme, the cost of the computation done on a single node in a single iteration is $C(b/K)$ and satisfies $C(b/K) \ge C(b)/K$ (the cost is sublinear in the batch size). 
In a system with no communication overhead and no overhead for summing the gradients, this approach could in principle achieve an accuracy of $a$ in time $N_a(b)C(b)/K$. 
This represents a linear speedup in the number of machines (for values of $K$ up to the batch size $b$). 

In practice, there are several important considerations. 
First, for the approximation $C(b/K) \approx C(b)/K$ to hold, $K$ must be much smaller than $b$, limiting the number of machines we can use to effectively parallelize the minibatch computation. 
One might imagine circumventing this limitation by using a larger batch size $b$. 
Unfortunately, the benefit of using larger batches is relatively modest. 
As the batch size $b$ increases, $N_a(b)$ does not decrease enough to justify the use of a very large value of $b$. 

Furthermore, the benefits of this approach depend greatly on the degree of communication overhead. 
If aggregating the gradients and broadcasting the model parameters requires $S$ units of time, then the time required by this approach is at least $C(b)/K + S$ per iteration and $N_a(b)(C(b)/K + S)$ to achieve an accuracy of $a$. 
Therefore, the maximum achievable speedup is $C(b)/(C(b)/K + S) \le C(b)/S$. 
We may expect $S$ to increase modestly as $K$ increases, but we suppress this effect here. 

\subsubsection{Limitations of SparkNet Parallelization}
\begin{figure}[t]
\centering
  \includegraphics[scale=1]{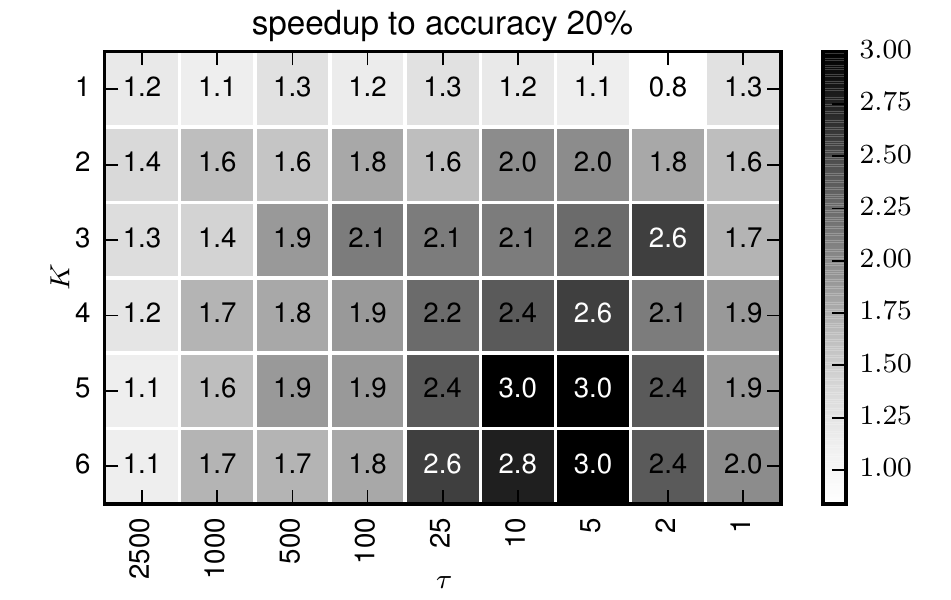}
  \caption{
    This figure shows the speedup $\tau M_a(b,\tau,K)/N_a(b)$ given by SparkNet's parallelization scheme relative to training on a single machine to obtain an accuracy of $a=20\%$. 
    Each grid square corresponds to a different choice of $K$ and $\tau$. 
    We show the speedup in the zero communication overhead setting. 
    This experiment uses a modified version of AlexNet on a subset of ImageNet ($100$ classes each with approximately $1000$ images). 
    Note that these numbers are dataset specific. 
    Nevertheless, the trends they capture are of interest. 
  }
  \label{fig:performance_heatmap2}
\end{figure}
The performance of the naive parallelization scheme is easily understood because its behavior is equivalent to that of the serial algorithm. 
In contrast, SparkNet uses a parallelization scheme that is not equivalent to serial SGD (described in \secref{sec:parallelizing_sgd}), and so its analysis is more complex. 

SparkNet's parallelization scheme proceeds in rounds (see \figref{fig:sparknet_flow}). 
In each round, each machine runs SGD for $\tau$ iterations with batch size $b$. 
Between rounds, the models on the workers are gathered together on the master, averaged, and broadcast to the workers. 

We use $M_a(b,K,\tau)$ to denote the number of rounds required to achieve an accuracy of $a$. 
The number of parallel iterations of SGD under SparkNet's parallelization scheme required to achieve an accuracy of $a$ is then $\tau M_a(b,K,\tau)$, and the wallclock time is 
\begin{equation} \label{eq:sparknet_time}
  (\tau C(b) + S) M_a(b,K,\tau) ,
\end{equation}
where $S$ is the time required to gather and broadcast model parameters. 

To measure the sensitivity of SparkNet's parallelization scheme to the parameters $\tau$ and $K$, we consider a grid of values of $K$ and $\tau$. 
For each pair of parameters, we run SparkNet using a modified version of AlexNet on a subset of ImageNet (the first $100$ classes each with approximately $1000$ data points) for a total of $20000$ parallel iterations. 
For each of these training runs, we compute the ratio $\tau M_a(b,K,\tau)/N_a(b)$. 
This is the speedup achieved relative to training on a single machine when $S=0$. 
In \figref{fig:performance_heatmap2}, we plot a heatmap of the speedup given by the SparkNet parallelization scheme under different values of $\tau$ and $K$. 

\figref{fig:performance_heatmap2} exhibits several trends. 
The top row of the heatmap corresponds to the case $K=1$, where we use only one worker. 
Since we do not have multiple workers to synchronize when $K=1$, the number of iterations $\tau$ between synchronizations does not matter, so all of the squares in the top row of the grid should behave similarly and should exhibit a speedup factor of $1$ (up to randomness in the optimization). 
The rightmost column of each heatmap corresponds to the case $\tau=1$, where we synchronize after every iteration of SGD. 
This is equivalent to running serial SGD with a batch size of $Kb$, where $b$ is the batchsize on each worker (in these experiments we use $b=100$). 
In this column, the speedup should increase sublinearly with $K$. 
We note that it is slightly surprising that the speedup does not increase monotonically from left to right as $\tau$ decreases. 
Intuitively, we might expect more synchronization to be strictly better (recall we are disregarding the overhead due to synchronization). 
However, our experiments suggest that modest delays between synchronizations can be beneficial. 

This experiment capture the speedup that we can expect from the SparkNet parallelization scheme in the case of zero communication overhead (the numbers are dataset specific, but the trends are of interest). 
Having measured these numbers, it is straightforward to compute the speedup that we can expect as a function of the communication overhead. 
\begin{figure}[t]
\centering
  \includegraphics[scale=1]{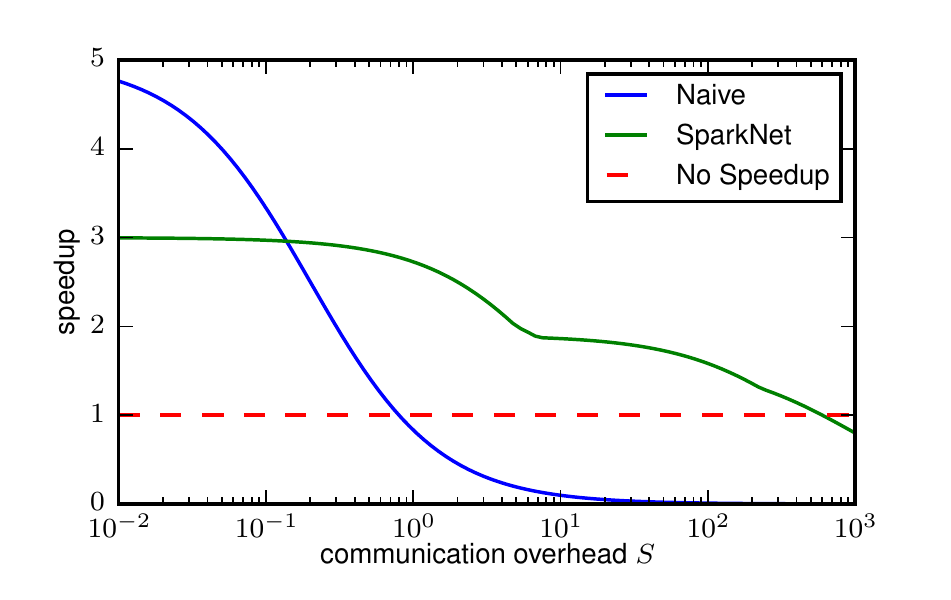}
  \caption{
    This figure shows the speedups obtained by the naive parallelization scheme and by SparkNet as a function of the cluster's communication overhead (normalized so that $C(b)=1$). 
    We consider $K=5$. 
    The data for this plot applies to training a modified version of AlexNet on a subset of ImageNet (approximately $1000$ images for each of the first $100$ classes). 
    The speedup obtained by the naive parallelization scheme is $C(b)/(C(b)/K + S)$. 
    The speedup obtained by SparkNet is $N_a(b) C(b) / [(\tau C(b) + S) M_a(b,K,\tau)]$ for a specific value of $\tau$. 
    The numerator is the time required by serial SGD to achieve an accuracy of $a$, and the denominator is the time required by SparkNet to achieve the same accuracy (see \eqref{eq:naive_time} and \eqref{eq:sparknet_time}). 
    For the optimal value of $\tau$, the speedup is $\max_{\tau} N_a(b) C(b) / [(\tau C(b) + S) M_a(b,K,\tau)]$. 
    To plot the SparkNet speedup curve, we maximize over the set of values $\tau \in \{1, 2, 5, 10, 25, 100, 500, 1000, 2500\}$ and use the values $M_a(b,K,\tau)$ and $N_a(b)$ from the experiments in the fifth row of \figref{fig:performance_heatmap2}. 
    In our experiments, we have $S \approx 20s$ and $C(b) \approx 2s$. 
  }
  \label{fig:parametric_speedup1}
\end{figure}

In \figref{fig:parametric_speedup1}, we plot the speedup expected both from naive parallelization and from SparkNet on a five-node cluster as a function of $S$ (normalized so that $C(b)=1$). 
As expected, naive parallelization gives a maximum speedup of $5$ (on a five-node cluster) when there is zero communication overhead (note that our plot does not go all the way to $S=0$), and it gives no speedup when the communication overhead is comparable to or greater than the cost of a minibatch computation. 
In contrast, SparkNet gives a relatively consistent speedup even when the communication overhead is $100$ times the cost of a minibatch computation. 

The speedup given by the naive parallelization scheme can be computed exactly and is given by $C(b)/(C(b)/K + S)$. 
This formula is essentially Amdahl's law. 
Note that when $S \ge C(b)$, the naive parallelization scheme is slower than the computation on a single machine. 
The speedup obtained by SparkNet is $N_a(b) C(b) / [(\tau C(b) + S) M_a(b,K,\tau)]$ for a specific value of $\tau$. 
The numerator is the time required by serial SGD to achieve an accuracy of $a$ from \eqref{eq:naive_time}, and the denominator is the time required by SparkNet to achieve the same accuracy from \eqref{eq:sparknet_time}. 
Choosing the optimal value of $\tau$ gives us a speedup of $\max_{\tau} N_a(b) C(b) / [(\tau C(b) + S) M_a(b,K,\tau)]$. 
In practice, choosing $\tau$ is not a difficult problem. 
The ratio $N_a(b)/(\tau M_a(b,K,\tau))$ (the speedup when $S=0$) degrades slowly as $\tau$ increases, so it suffices to choose $\tau$ to be a small multiple of $S$ (say $5S$) so that the algorithm spends only a fraction of its time in communication. 

When plotting the SparkNet speedup in \figref{fig:parametric_speedup1}, we do not maximize over all positive integer values of $\tau$ but rather over the set $\tau \in \{1, 2, 5, 10, 25, 100, 500, 1000, 2500\}$, and we use the values of $N_a(b)$ and $M_a(b,K,\tau)$ corresponding to the fifth row of \figref{fig:performance_heatmap2}. 
Including more values of $\tau$ would only increase the SparkNet speedup. 
The distributed training of deep networks is typically thought of as a communication-intensive procedure. 
However, \figref{fig:parametric_speedup1} demonstrates the value of SparkNet's parallelization scheme even in the most bandwidth-limited settings. 

The naive parallelization scheme may appear to be a straw man. 
However, it is a frequently-used approach to parallelizing SGD \citep{noel2015large,iandola2015firecaffe}, especially when asynchronous updates are not an option (as in computational frameworks like MapReduce and Spark). 

\subsection{Training Benchmarks} \label{sec:training_benchmarks}

\begin{figure}
\centering
\begin{minipage}[t]{.47\textwidth}
  \centering
  \includegraphics{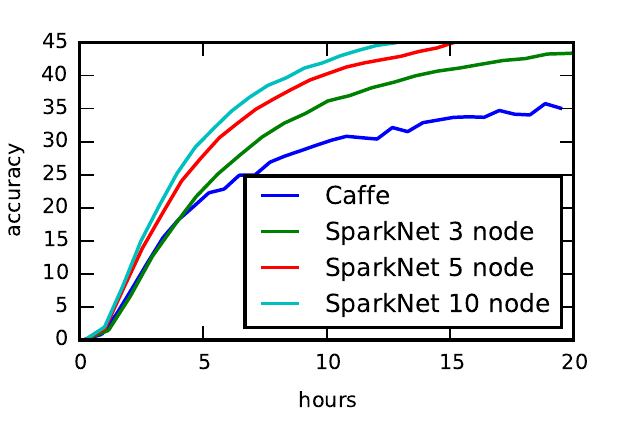}
  \caption{
    This figure shows the performance of SparkNet on a $3$-node, $5$-node, and $10$-node cluster, where each node has $1$ GPU. 
    In these experiments, we use $\tau=50$. 
    The baseline was obtained by running Caffe on a single GPU with no communication. 
    The experiments are performed on ImageNet using AlexNet. 
}
  \label{fig:scaling_experiments}
\end{minipage}%
\hspace{15pt}
\begin{minipage}[t]{.47\textwidth}
  \centering
  \includegraphics{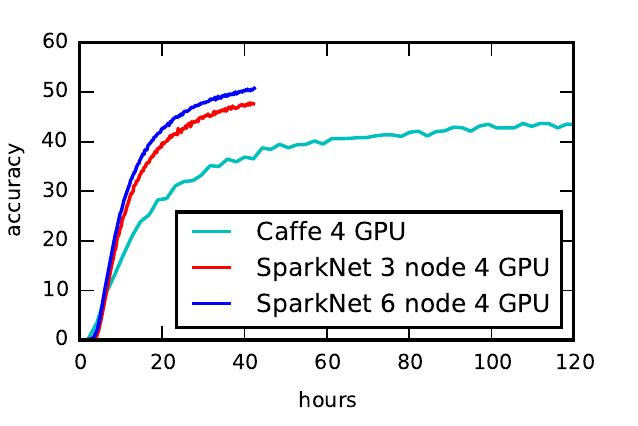}
  \caption{
    This figure shows the performance of SparkNet on a 3-node cluster and on a 6-node cluster, where each node has 4 GPUs. 
    In these experiments, we use $\tau=50$. 
    The baseline uses Caffe on a single node with 4 GPUs and no communication overhead. 
    The experiments are performed on ImageNet using GoogLeNet.
}
  \label{fig:googlenet_experiments}
\end{minipage}
\end{figure}

To explore the scaling behavior of our algorithm and implementation, we perform experiments on EC2 using clusters of g2.8xlarge nodes. 
Each node has four NVIDIA GRID GPUs and 60GB memory. 
We train the default Caffe model of AlexNet \citep{krizhevsky2012imagenet} on the ImageNet dataset \citep{russakovsky2015imagenet}. 
We run SparkNet with $K=3$, $5$, and $10$ and plot the results in \figref{fig:scaling_experiments}. 
For comparison, we also run Caffe on the same cluster with a single GPU and no communication overhead to obtain the $K=1$ plot. 
These experiments use only a single GPU on each node. 
To measure the speedup, we compare the wall-clock time required to obtain an accuracy of $45\%$. 
With $1$ GPU and no communication overhead, this takes $55.6$ hours. 
With $3$, $5$, and $10$ GPUs, SparkNet takes $22.9$, $14.5$, and $12.8$ hours, giving speedups of $2.4$, $3.8$, and $4.4$. 

We also train the default Caffe model of GoogLeNet \citep{szegedy2015going} on ImageNet. 
We run SparkNet with $K=3$ and $K=6$ and plot the results in \figref{fig:googlenet_experiments}. 
In these experiments, we use Caffe's multi-GPU support to take advantage of all four GPUs within each node, and we use SparkNet's parallelization scheme to handle parallelism across nodes. 
For comparison, we train Caffe on a single node with four GPUs and no communication overhead. 
To measure the speedup, we compare the wall-clock time required to obtain an accuracy of $40\%$. 
Relative to the baseline of Caffe with four GPUs, SparkNet on 3 and 6 nodes gives speedups of $2.7$ and $3.2$. 
Note that this is on top of the speedup of roughly $3.5$ that Caffe with four GPUs gets over Caffe with one GPU, so the speedups that SparkNet obtains over Caffe on a single GPU are roughly $9.4$ and $11.2$. 

\begin{figure}
  \centering
  \includegraphics{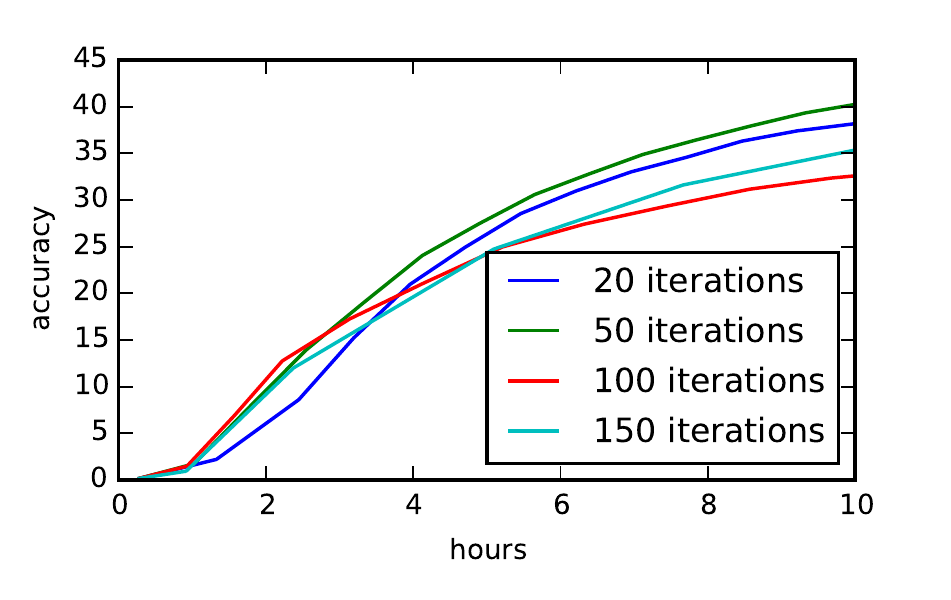}
  \caption{
    This figure shows the dependence of the parallelization scheme described in \secref{sec:parallelizing_sgd} on $\tau$. 
    Each experiment was run with $K=5$ workers. 
    This figure shows that good performance can be achieved without collecting and broadcasting the model after every SGD update.
  }
  \label{fig:n_sync_experiments}
\end{figure}

Furthermore, we explore the dependence of the parallelization scheme described in \secref{sec:parallelizing_sgd} on the parameter $\tau$ which determines the number of iterations of SGD that each worker does before synchronizing with the other workers. 
These results are shown in \figref{fig:n_sync_experiments}. 
Note that in the presence of stragglers, it suffices to replace the fixed number of iterations $\tau$ with a fixed length of time, but in our experimental setup, the timing was sufficiently consistent and stragglers did not arise. 
The single GPU experiment in \figref{fig:scaling_experiments} was trained on a single GPU node with no communication overhead.

\section{Related Work}
Much work has been done to build distributed frameworks for training deep networks. 
\citet{coates2013deep} build a model-parallel system for training deep networks on a GPU cluster using MPI over Infiniband. 
\citet{dean2012large} build DistBelief, a distributed system capable of training deep networks on thousands of machines using stochastic and batch optimization procedures. 
In particular, they highlight asynchronous SGD and batch L-BFGS. 
Distbelief exploits both data parallelism and model parallelism. 
\citet{chilimbi2014project} build Project Adam, a system for training deep networks on hundreds of machines using asynchronous SGD. 
\citet{li2014scaling,ho2013more} build parameter servers to exploit model and data parallelism, and though their systems are better suited to sparse gradient updates, they could very well be applied to the distributed training of deep networks. 
More recently, \citet{abadi2015tensorflow} build TensorFlow, a sophisticated system for training deep networks and more generally for specifying computation graphs and performing automatic differentiation. 
\citet{iandola2015firecaffe} build FireCaffe, a data-parallel system that achieves impressive scaling using naive parallelization in the high-performance computing setting. 
They minimize communication overhead by using a tree reduce for aggregating gradients in a supercomputer with Cray Gemini interconnects. 

These custom systems have numerous advantages including high performance, fine-grained control over scheduling and task placement, and the ability to take advantage of low-latency communication between machines. 
On the other hand, due to their demanding communication requirements, they are unlikely to exhibit the same scaling on an EC2 cluster. 
Furthermore, due to their nature as custom systems, they lack the benefits of tight integration with general-purpose computational frameworks such as Spark. 
For some of these systems, preprocessing must be done separately by a MapReduce style framework, and data is written to disk between segments of the pipeline. 
With SparkNet, preprocessing and training are both done in Spark. 

Training a machine learning model such as a deep network is often one step of many in real-world data analytics pipelines \citep{sparks2015keystoneml}. 
Obtaining, cleaning, and preprocessing the data are often expensive operations, as is transferring data between systems. 
Training data for a machine learning model may be derived from a streaming source, from a SQL query, or from a graph computation. 
A user wishing to train a deep network in a custom system on the output of a SQL query would need a separate SQL engine. 
In SparkNet, training a deep network on the output of a SQL query, or a graph computation, or a streaming data source is straightforward due to its general purpose nature and its support for SQL, graph computations, and data streams \citep{armbrust2015spark,gonzalez2014graphx,zaharia2013discretized}. 

Some attempts have been made to train deep networks in general-purpose computational frameworks, however, existing work typically hinges on extremely low-latency intra-cluster communication. 
\citet{noel2015large} train deep networks in Spark on top of YARN using SGD and leverage cluster resources to parallelize the computation of the gradient over each minibatch. 
To achieve competitive performance, they use remote direct memory accesses over Infiniband to exchange model parameters quickly between GPUs. 
In contrast, SparkNet tolerates low-bandwidth intra-cluster communication and works out of the box on Amazon EC2. 

A separate line of work addresses speeding up the training of deep networks using single-machine parallelism. 
For example, Caffe con Troll \citep{abuzaid2015caffe} modifies Caffe to leverage both CPU and GPU resources within a single node. 
These approaches are compatible with SparkNet and the two can be used in conjunction. 

Many popular computational frameworks provide support for training machine learning models \citep{meng2015mllib} such as linear models and matrix factorization models. 
However, due to the demanding communication requirements and the larger scale of many deep learning problems, these libraries have not been extended to include deep networks. 

Various authors have studied the theory of averaging separate runs of SGD. 
In the bandwidth-limited setting, \citet{zinkevich2010parallelized} analyze a simple algorithm for convex optimization that is easily implemented in the MapReduce framework and can tolerate high-latency communication between machines. 
\citet{sixin2015deep} define a parallelization scheme that penalizes divergences between parallel workers, and they provide an analysis in the convex case. 
\citet{zhang2015splash} propose a general abstraction for parallelizing stochastic optimization algorithms along with a Spark implementation. 

\section{Discussion}
We have described an approach to distributing the training of deep networks in communication-limited environments that lends itself to an implementation in batch computational frameworks like MapReduce and Spark. 
We provide SparkNet, an easy-to-use deep learning implementation for Spark that is based on Caffe and enables the easy parallelization of existing Caffe models with minimal modification. 
As machine learning increasingly depends on larger and larger datasets, integration with a fast and general engine for big data processing such as Spark allows researchers and practitioners to draw from a rich ecosystem of tools to develop and deploy their models. 
They can build models that incorporate features from a variety of data sources like images on a distributed file system, results from a SQL query or graph database query, or streaming data sources. 

Using a smaller version of the ImageNet benchmark we quantify the speedup achieved by SparkNet as a function of the size of the cluster, the communication frequency, and the cluster's communication overhead. 
We demonstrate that our approach is effective even in highly bandwidth-limited settings. 
On the full ImageNet benchmark we showed that our system achieves a sizable speedup over a single node experiment even with few GPUs. 

The code for SparkNet is available at \url{https://github.com/amplab/SparkNet}. 
We invite contributions and hope that the project will help bring a diverse set of deep learning applications to the Spark community.

\subsubsection*{Acknowledgments}

We would like to thank Cyprien Noel, Andy Feng, Tomer Kaftan, Evan Sparks, and Shivaram Venkataraman for valuable advice. 
This research is supported in part by NSF grant number DGE-1106400. 
This research is supported in part by NSF CISE Expeditions Award CCF-1139158, DOE Award SN10040 DE-SC0012463, and DARPA XData Award FA8750-12-2-0331, and gifts from Amazon Web Services, Google, IBM, SAP, The Thomas and Stacey Siebel Foundation, Adatao, Adobe, Apple, Blue Goji, Bosch, Cisco, Cray, Cloudera, EMC2, Ericsson, Facebook, Fujitsu, Guavus, HP, Huawei, Informatica, Intel, Microsoft, NetApp, Pivotal, Samsung, Schlumberger, Splunk, Virdata and VMware.

\bibliography{refs}
\bibliographystyle{iclr2016_conference}

\end{document}